\documentclass[final,5p,times]{elsarticle}

\usepackage{amssymb}
\usepackage{epsfig}
\usepackage{amsmath}
\usepackage{amsfonts}
\usepackage{bm}
\usepackage{graphicx}
\usepackage{appendix}
\usepackage{mathtools}
\usepackage{caption}
\usepackage{color}
\usepackage[subrefformat=parens]{subcaption}
\usepackage[normalem]{ulem}
\usepackage{url}
\usepackage{multicol}
\usepackage{microtype}
\usepackage{booktabs} 
\usepackage{wrapfig}
\usepackage {threeparttable}
\usepackage{booktabs} 
\usepackage{multirow} 
\usepackage{arydshln} 
\usepackage{array}
\newcolumntype{C}[1]{>{\centering\arraybackslash}m{#1}} 

\journal{Geodata and AI}

\begin{document}

\begin{frontmatter}


\author{Taiga Saito\corref{cor1}\fnref{label1}}
\ead{taiga.saito.r3@dc.tohoku.ac.jp}
\cortext[cor1]{Corresponding author}

\title{Tabular foundation model for GEOAI benchmark problems BM/AirportSoilProperties/2/2025}


\author[label1]{Yu Otake}
\author[label2,label3]{Stephen Wu}

\affiliation[label1]{
    organization={Department of Civil and Environmental Engineering, Tohoku University},
    addressline={6-6-06 Aramaki Aoba, Aoba-ku},
    city={Sendai},
    state={Miyagi},
    postcode={980-8579},
    country={Japan}
}
 \affiliation[label2]{
     organization={Research Organization of Information and Systems, The Institute of Statistical Mathematics},
     addressline={10-3 Midori-cho},
     city={Tachikawa},
     state={Tokyo},
     postcode={190-8562},
     country={Japan}
 }
 \affiliation[label3]{
     organization={The Graduate University for Advanced Studies, Department of Statistical Science},
     addressline={10-3 Midori-cho},
     city={Tachikawa},
     state={Tokyo},
     postcode={190-8562},
     country={Japan}
 }

\begin{abstract}
This paper presents a novel application of the Tabular Prior-Data Fitted Network (TabPFN) -- a transformer-based foundation model for tabular data -- to geotechnical site characterization problems defined in the GEOAI benchmark BM/AirportSoilProperties/2/2025. Two tasks are addressed: (1) predicting the spatial variation of undrained shear strength ($s_u$) across borehole depth profiles, and (2) imputing missing mechanical parameters in a dense-site dataset. We apply TabPFN in a zero-training, few-shot, in-context learning setting—without hyper-parameter tuning—and provide it with additional context from the big indirect database (BID).
The study demonstrates that TabPFN, as a general-purpose foundation model, achieved superior accuracy and well-calibrated predictive distributions compared to a conventional hierarchical Bayesian model (HBM) baseline, while also offering significant gains in inference efficiency. In Benchmark Problem \#1 (spatial $s_u$ prediction), TabPFN outperformed the HBM in prediction accuracy and delivered an order-of-magnitude faster runtime. In Benchmark Problem \#2 (missing mechanical parameter imputation), TabPFN likewise achieved lower RMSE for all target parameters with well-quantified uncertainties, though its cumulative computation cost was higher than HBM’s due to its one-variable-at-a-time inference. These results mark the first successful use of a tabular foundation model in geotechnical modeling, suggesting a potential paradigm shift in probabilistic site characterization.
\end{abstract}



\begin{keyword}
TabPFN\sep Foundation Models\sep Transformer\sep Soil property prediction\sep Probabilistic site characterization
\end{keyword}

\end{frontmatter}



\section{Introduction}

Probabilistic site characterization, a cornerstone of modern geotechnical design, aims to predict subsurface soil properties by integrating sparse site-specific data with broader big indirect database (BID) \cite{PHOON2022967}. 
Otake et al. \cite{OTAKE2025100012} recently introduced the GEOAI benchmark BM/AirportSoilProperties/2/2025 to formalize this challenge via standardized tasks of spatial prediction and data imputation.
Conventional approaches center on hierarchical Bayesian models (HBMs), which provide a rigorous framework for encoding domain knowledge and combining data sources \cite{Ching2021hbm, BOZORGZADEH20191056}. 
HBMs have been successfully applied by many researchers to tackle the ``site recognition challenge''  \cite{Phoon02012022, doi:10.1061/AJRUA6.RUENG-1553}. 
However, HBM requires significant modeling effort and can be computationally intensive, which has motivated the exploration of more direct, data-driven alternatives \cite{WU2022102253, Sharma2022, SHARMA2023105624, CAI2024107537, CAI2025107072, SAITO2025100009}.

A promising alternative arises from a concurrent paradigm shift in artificial intelligence, catalyzed by the introduction of the Transformer architecture \cite{NIPS2017_3f5ee243}. 
The Transformer’s core innovation—the self-attention mechanism—dispensed with the sequential processing of recurrent neural networks, enabling massive parallelization and the scaling of models to unprecedented sizes. This architectural breakthrough paved the way for the rise of ``foundation models''—large models pre-trained on vast quantities of data that can be adapted to a wide range of downstream tasks \cite{NIPS2017_3f5ee243}. As described by Bommasani et al. \cite{Bommasani2021}, these models are characterized by ``emergence'' (new capabilities arising implicitly from scale) and ``homogenization'' (a single powerful model serving many diverse applications).

This new paradigm was popularized by large language models (LLMs) like GPT-4 \cite{achiam2023gpt}. The seminal work by Brown et al. \cite{NEURIPS2020_1457c0d6} demonstrated that at a sufficient scale, models such as GPT-3 acquire a remarkable capability known as ``in-context learning.'' This is the ability to perform novel tasks based solely on a few examples (few-shot) or instructions provided in the input, without any updates to the model’s weights (i.e. no task-specific fine-tuning). This in-context learning mechanism represents a fundamental departure from the traditional supervised learning pipeline. Notably, this paradigm is beginning to find applications in specialized fields such as geotechnical engineering. Researchers are exploring the use of LLMs to streamline workflows, from interpreting geological reports to assisting in design calculations \cite{WU2024101471, Wu03042025, fan2025domainadaptationlargelanguage}. Notably, Wu et al. \cite{Wu03042025} demonstrated that integrating LLMs into geotechnical practice can enhance efficiency, data processing, and decision-making, suggesting a move towards more integrated, data-driven approaches in the discipline.

The Tabular Prior-Data Fitted Network (TabPFN) is a direct application of these principles to the domain of tabular data \cite{hollmann2023tabpfntransformersolvessmall}. 
TabPFN is a Transformer-based foundation model tailored for tabular prediction tasks.
It operates as a Prior-Data Fitted Network (PFN), meaning it is not trained on user-provided data. Instead, it is pre-trained a single time on millions of synthetic datasets generated from a broad prior over structural causal models. By pre-training on millions of synthetic datasets, TabPFN learns to approximate Bayesian inference in a single forward pass. This allows it to function as a ``few-shot'' predictor, requiring no hyperparameter tuning and offering a significant speedup over conventional methods \cite{Hollmann:2025aa}.

This paper presents the first application of TabPFN in geotechnical engineering, using the GEOAI benchmark \cite{OTAKE2025100012} to test a central hypothesis: Can a ``generalist'' data-driven model, without any explicit geotechnical theory, match or exceed the performance of a ``specialist'' HBM built on domain knowledge? A positive answer would not only provide a powerful new tool for practitioners but also signal a potential paradigm shift in how we approach site characterization, mirroring trends seen in fields like computer vision and natural language processing \cite{NIPS2017_3f5ee243}. This study provides quantitative evidence of TabPFN’s superior predictive accuracy and efficiency, and introduces the concept of ``geotechnical prompt engineering'' as a practical paradigm for integrating domain knowledge into a data-centric workflow.

\section{Methodology} 

\subsection{Benchmark Data}

The case study focuses on a large offshore airport underlain by soft clay of the Yurakucho Formation. A comprehensive soil investigation at this site yielded the ``Tokyo-CLAY/14/67760'' database \cite{Ishii_1985, Watabe_2004, Otake2024, SAITO2025106826}, comprising 2,922 boreholes and 67,760 records over a 10 km × 10 km area. From this, a smaller ``verification site'' (300 m × 300 m) with denser sampling (51 boreholes, 1,001 records; Otake et al. \cite{OTAKE2025100012}) was selected for detailed analysis.

At each location, eleven geotechnical parameters were measured and split into two groups:
\begin{itemize}
  \item \textbf{Index properties}—economical, frequently measured: degree of saturation $S_r$, total unit weight $\gamma_t$, void ratio $e$, liquid limit $\mathrm{LL}$, plastic limit $\mathrm{PL}$, and water content $w$.
  \item \textbf{Mechanical properties}—costly, less frequent: undrained shear strength $s_u$ (unconfined compression), undrained secant modulus $E_u$, preconsolidation stress $\sigma'_p$, compression index $C_c$, and coefficient of consolidation $c_v$.
\end{itemize}

Otake et al. \cite{OTAKE2025100012} define two benchmark prediction tasks using these data:

\begin{enumerate}
  \item \textbf{Benchmark Problem \#1: Predicting the spatial variation of $s_u$.}  
    For five selected boreholes (B1–B5) in the verification site, predict the full depth profile of  $s_u$ and compare to the observed values. Performance is evaluated by the RMSE of the predicted mean profile and the coverage of 95\% prediction intervals at each depth.

  \item \textbf{Benchmark Problem \#2: Predicting missing mechanical parameters.}  
    Within the full Tokyo-CLAY/14/67760 database, 20 records from the verification site that lack some or all mechanical property measurements are identified. The goal is to predict all five mechanical parameters ($s_u, E_u, \sigma'_p, C_c, c_v$) for each of these incomplete records. Evaluation uses RMSE and 95\% interval coverage for the predictions.
\end{enumerate}

In both tasks, incorporating big indirect database (BID)—large, non–site-specific datasets that supply statistical context—is crucial \cite{PHOON2022967, Phoon02012022}. In our implementation, the BID sources include: (1) the large-scale offshore airport clay database in Japan (Local-BID, Tokyo-CLAY/14/67760), (2) spatial subsets of this database (Local-BID-V or Cluster-BID; see Otake et al. \cite{OTAKE2025100012}), and (3) other global databases (Global-BID, e.g. CLAY/10/7490 developed by Ching and Phoon \cite{doi:10.1139/cgj-2013-0262}).

A key question is how to leverage these BIDs effectively under different modeling frameworks. The HBM uses BID-derived statistics (e.g. the mean and variance of $s_u$, and inter-parameter correlations) to inform its prior distributions. In contrast, TabPFN treats BID data simply as additional ``context'' examples provided alongside the site-specific data, as described next.

\subsection{The TabPFN Model and In-Context Learning Approach}

\begin{figure*}[htb]
\centering
\includegraphics[width=\linewidth]{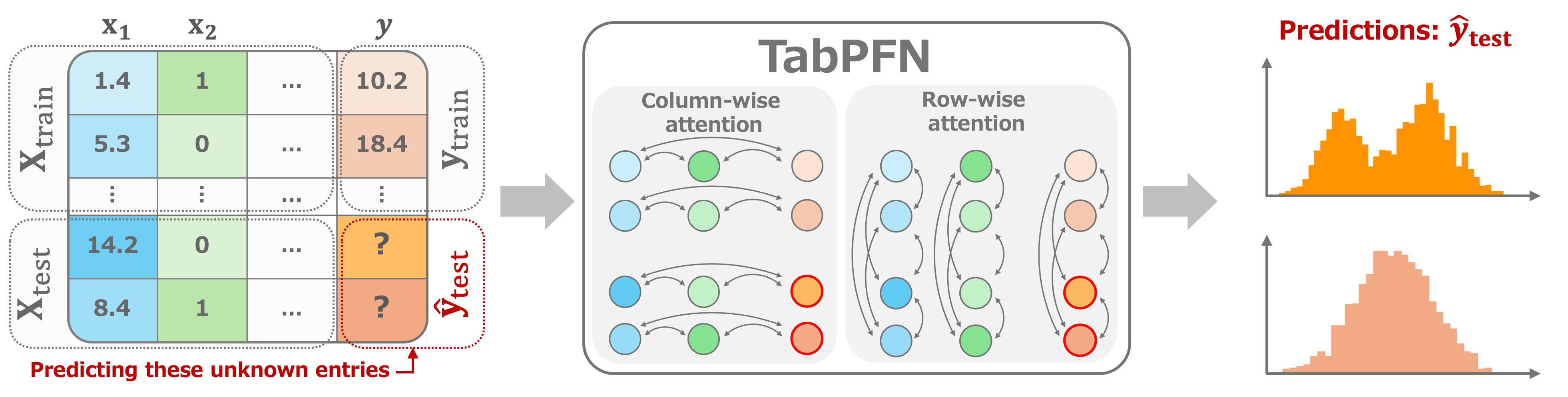}
\caption{Schematic illustration of the TabPFN architecture. TabPFN uses a transformer encoder with a two-dimensional attention mechanism adapted for tabular data, where each cell attends across its row and column. This enables the model to capture interactions among features and samples efficiently, with invariance to data ordering. The model is pre-trained on over 100 million synthetic tabular tasks, learning a generic algorithm for prediction. At inference, it accepts a new dataset’s training samples and features as input context and produces predictions for target outputs in a single forward pass (no gradient updates). The output is a calibrated predictive distribution over target values, rather than a single point estimate, reflecting the model’s approximation of Bayesian posterior uncertainty.}
\label{fig:tabpfn_fw}
\end{figure*}

TabPFN is a Transformer-based architecture specifically designed for tabular data \cite{hollmann2023tabpfntransformersolvessmall, Hollmann:2025aa} and exploits the principle of in-context learning \cite{NEURIPS2020_1457c0d6}. As shown in Fig.\ref{fig:tabpfn_fw}, its architecture diverges from standard sequence models by assigning a unique representation to each table cell. It then employs a two-way attention mechanism: each cell attends to other features within the same sample (row-wise attention) while simultaneously attending to other samples within the same feature (column-wise attention). This design makes the model invariant to permutations of both rows (samples) and columns (features) — a critical property for general-purpose tabular data analysis.

As a Prior-Fitted Network, TabPFN is not trained on the user’s data. Instead, it is pre-trained on a vast number of synthetic datasets generated from a broad prior over structural causal models, thereby learning to mimic Bayesian posterior-predictive inference. At inference time, the entire training set $(\mathbf{X}_{\text{train}}, \mathbf{y}_{\text{train}})$ is combined with the test inputs $\mathbf{X}_{\text{test}}$ and serialized into one long context sequence. In a single forward pass — without any gradient updates — the model outputs an approximation to the posterior predictive distribution:

\[
p\bigl(\,\hat{\mathbf{y}}_{\text{test}}\mid \mathbf{X}_{\text{test}},\,\mathbf{X}_{\text{train}},\,\mathbf{y}_{\text{train}}\,\bigr)
\]

For regression tasks, TabPFN represents this distribution as a piecewise-constant function, enabling it to capture complex uncertainties (including multi-modal outcomes). Consequently, TabPFN can model intricate relationships within the data without requiring explicit user-specified modeling.

\subsection{Problem Setup for TabPFN}

\begin{figure*}[htb]
\centering
\includegraphics[width=1\linewidth]{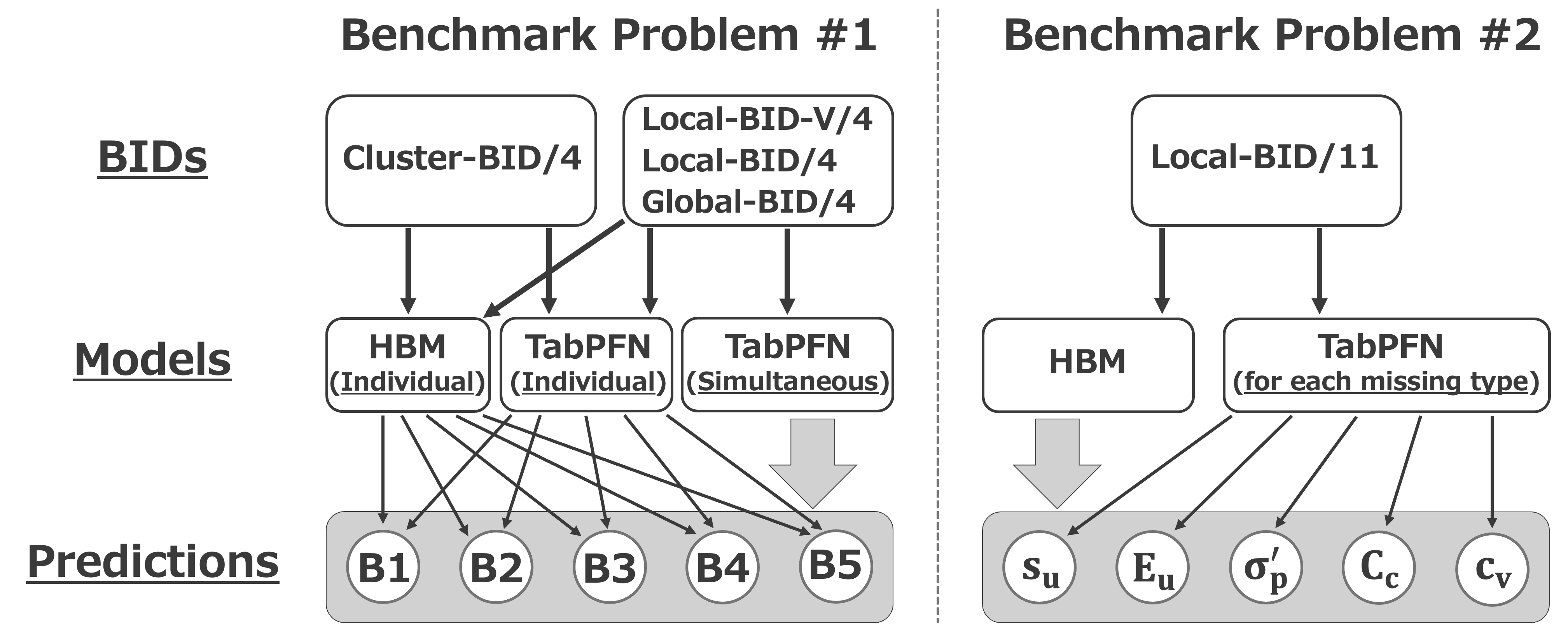}
\caption{Overview of the inference workflows evaluated in this study. For Benchmark Problem \#1, four BIDs are tested: Cluster‑BID/4, Local‑BID‑V/4, Local‑BID/4 and Global‑BID/4.  
Each BID is supplied to (i) a hierarchical Bayesian model (HBM) trained individually for every target borehole, (ii) TabPFN run in the same individual‑borehole setting, and (iii) TabPFN run once in a simultaneous setting that predicts the five verification boreholes (B1–B5) in a single forward pass.  The arrows indicate which BID–model pairs produce each prediction.  
For Benchmark Problem \#2, the Local‑BID/11 database is combined with the verification‑site table that contains up to five missing mechanical parameters—\(s_u\), \(E_u\), \(\sigma'_p\), \(C_c\) and \(c_v\). For each missing‑data type, TabPFN is invoked once per target variable, while the baseline HBM jointly models all variables.  Arrows show how the two model families contribute to the imputation of each parameter type.}
\label{fig:benchmark_setting}
\end{figure*}

To apply TabPFN to Benchmark Problem \#1, we devised two in-context learning scenarios — both using few-shot inference (no weight fine-tuning) — to evaluate the predictive utility of each BID source, as shown on the left side of Fig.\ref{fig:benchmark_setting}:

\begin{enumerate}
  \item \textbf{Individual Borehole Prediction.} 
    In this scenario, we assess each BID’s ability to predict a single borehole’s profile. To ensure comparability, this setting was designed to mirror the HBM framework of Otake et al. \cite{OTAKE2025100012}. For each of the four BIDs and each of the five target boreholes (B1–B5), we construct a unique context as follows:

    \begin{itemize}
        \item \textbf{Training set} $(\mathbf{X}_{\text{train}}, \mathbf{y}_{\text{train}})$\textbf{:} all records from the selected BID are combined with all records from the target borehole where $s_u$ was measured.
        \item \textbf{Test set} $(\mathbf{X}_{\text{test}})$\textbf{:} all records from the target borehole where $s_u$ was not measured. TabPFN is then prompted to predict these missing $s_u$ values.
    \end{itemize}

    This process is repeated for all 20 combinations (4 BIDs $\times$ 5 boreholes), allowing a direct comparison of each BID’s effectiveness as predictive context.

  \item \textbf{Simultaneous All-Borehole Prediction.} 
    In this scenario, we evaluate each BID’s utility for a comprehensive site-wide prediction task. The context is constructed by combining: (i) all records from the selected BID, and (ii) the records from all five site boreholes where $s_u$ was measured.

    \begin{itemize}
        \item \textbf{Training set} $(\mathbf{X}_{\text{train}}, \mathbf{y}_{\text{train}})$\textbf{:} includes all BID records plus the combined measured $s_u$ records from boreholes B1–B5.
        \item \textbf{Test set} $(\mathbf{X}_{\text{test}})$\textbf{:} all records from B1–B5 where $s_u$ was not measured. TabPFN is prompted to predict the missing $s_u$ values for all boreholes simultaneously.
    \end{itemize}

    We conducted this scenario for three of the BIDs (Local-BID-V/4, Local-BID/4, and Global-BID/4). The Cluster-BID/4 was excluded, because its methodology uses different database subsets for each borehole, making a single site-wide context infeasible for comparison.
\end{enumerate}

For Benchmark Problem \#2, TabPFN was applied to predict the missing mechanical parameters using an in-context learning approach that leveraged the Local-BID/11 database as the sole source of contextual information, as shown on the right side of Fig.\ref{fig:benchmark_setting}. A key aspect of this benchmark is that multiple mechanical parameters (up to five types) can be missing in various combinations. The dataset defines four distinct missingness patterns, which we address independently. Since TabPFN is designed to predict a single target variable at a time, we cannot impute all missing values simultaneously. Instead, we construct a separate prediction task for each target parameter within each missingness pattern.

For a given target parameter (e.g., $E_u$) and a specific missingness pattern, the TabPFN context is constructed as follows:

\begin{itemize}
    \item \textbf{Training set} $(\mathbf{X}_{\text{train}}, \mathbf{y}_{\text{train}})$\textbf{:} all records from the Local-BID/11 database, combined with any records from the problem dataset (Otake et al. \cite{OTAKE2025100012}, Table 1) where the target parameter (e.g. $E_u$) is known.
    \item \textbf{Test set} $(\mathbf{X}_{\text{test}})$\textbf{:} the records corresponding to the chosen missingness pattern in which the target parameter is missing. The features in $(\mathbf{X}_{\text{test}})$ consist of whatever parameters remain (i.e. the non-missing measurements for those records)
\end{itemize}

Using this approach, we build a unique model setup for each combination of missingness pattern and target parameter. In total, 14 distinct TabPFN prediction models were constructed to cover all cases, enabling a detailed assessment of TabPFN’s imputation capabilities across different scenarios.

To implement the above methods, we used the open-source TabPFN code (PriorLabs \cite{tabpfn_github}) in Python, leaving the model in its default pre-trained state. Input tables were prepared for each scenario, with categorical identifiers (e.g. borehole or site IDs) integer-encoded. Because $s_u$ and the other target variables are continuous, TabPFN outputs a discrete approximation of the posterior predictive distribution. From this distribution, we extracted the mean predictions (used to compute RMSE) and relevant quantiles (used to evaluate prediction interval coverage). For our HBM baseline, we followed the implementation and results reported by Otake et al. \cite{OTAKE2025100012}, and we compare TabPFN’s performance to those HBM benchmarks under each task and BID scenario.

\subsection{Evaluation Metrics and Baseline}
We evaluate the models on two primary criteria, with the HBM results from Otake et al. \cite{OTAKE2025100012} serving as the performance baseline:

\begin{enumerate}
    \item \textbf{Predictive Accuracy:} Measured by the Root Mean Squared Error (RMSE) between the model’s predicted mean and the true values. For Benchmark \#1, an RMSE is computed for each borehole’s $s_u$ profile. For Benchmark \#2, RMSE is computed for each imputed mechanical parameter.
    
    \item \textbf{Computational Cost:} Measured by wall-clock runtime. For TabPFN, this is the inference time of a forward pass only (since pre-training is a one-time sunk cost). For HBM, it includes the time for both model training on the BID and subsequent inference. TabPFN runtimes were measured on an Apple M1 Pro CPU with 16 GB RAM, while HBM runtimes were taken from the literature \cite{OTAKE2025100012} for consistency.
\end{enumerate}

\section{Results}

\subsection{Benchmark Problem \#1: Predicting the spatial variation of $s_u$}

In Benchmark Problem \#1, TabPFN consistently demonstrated superior performance over the HBM baseline across all key metrics: predictive accuracy, uncertainty calibration, and computational efficiency.

\begin{figure*}[htb]
\centering
\includegraphics[width=\linewidth]{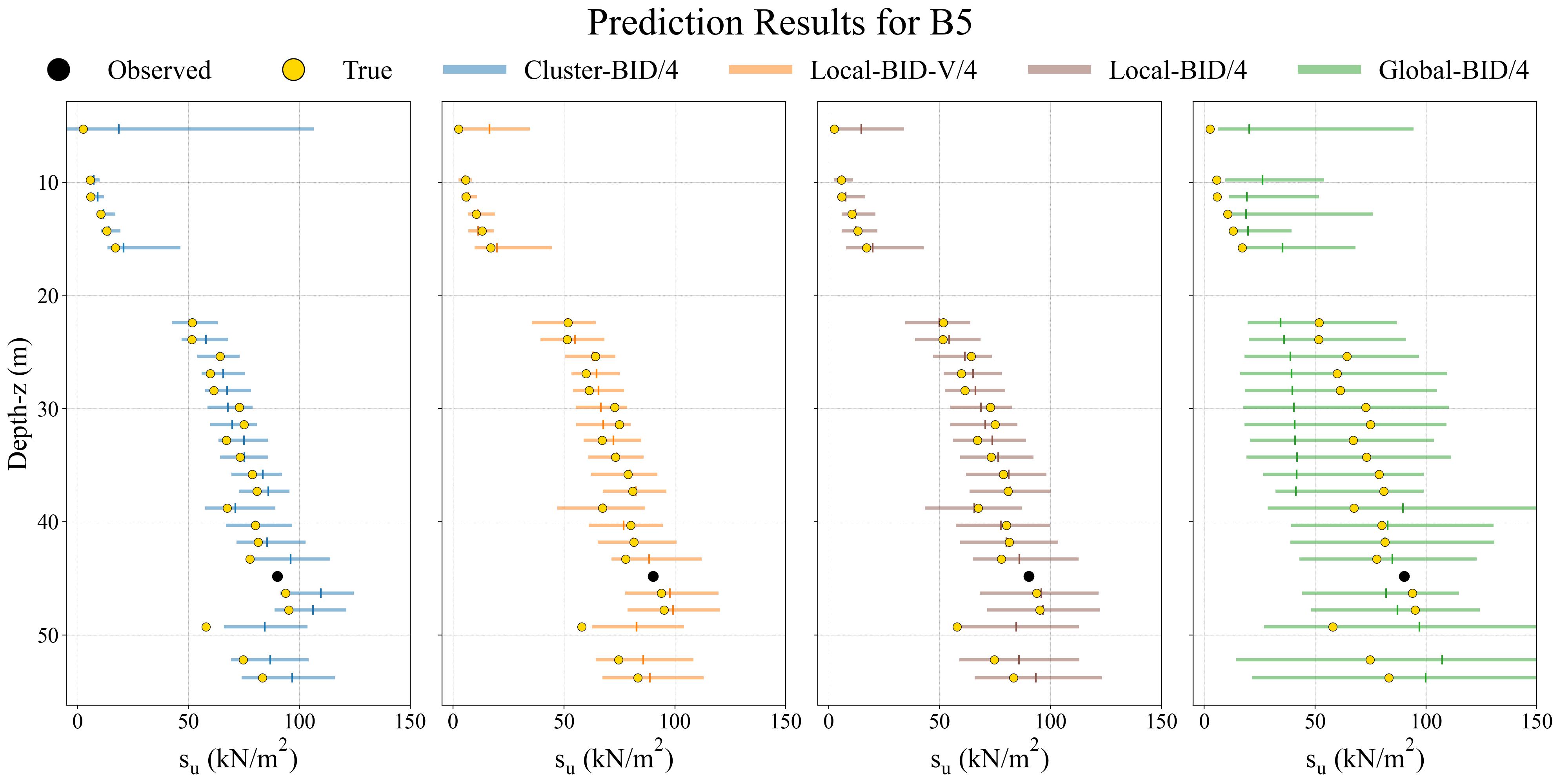}
\caption{Prediction results for the $s_u$ of borehole B5. Each panel shows the TabPFN predictions using a different BID as the predictive context. Observed $s_u$ values provided to the model are shown as black circles. The colored points and horizontal bars represent the mean and the 95\% credible interval of TabPFN's posterior predictive distribution, respectively.}
\label{fig:c1_pred_dist_B5}
\end{figure*}

\begin{figure*}[htb]
\centering
\includegraphics[width=\linewidth]{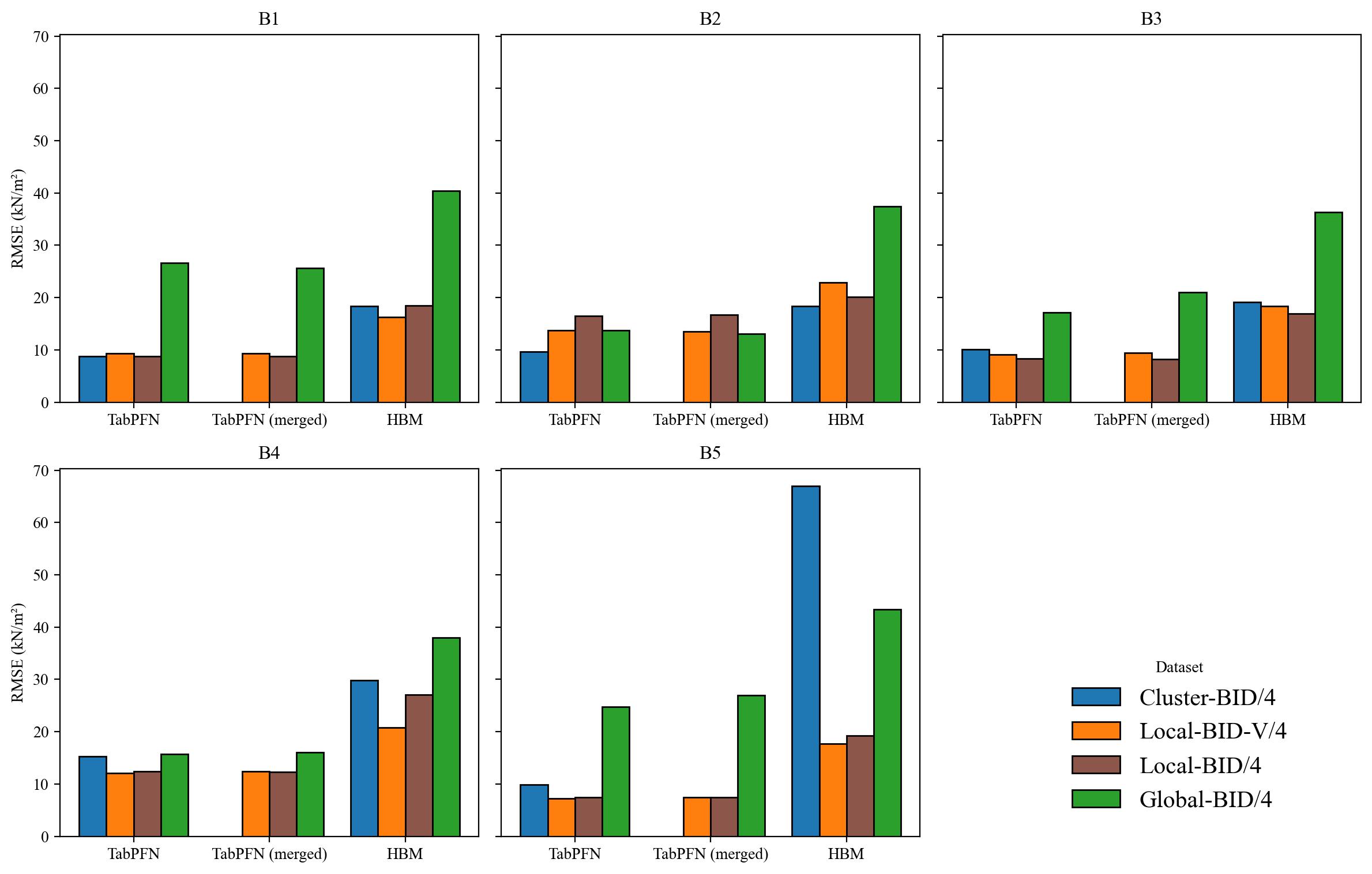}
\caption{Comparison of RMSE for $s_u$ predictions in Benchmark Problem \#1. Each barplot (B1–B5) compares the performance of three methods: the HBM baseline, TabPFN with individual predictions (``TabPFN''), and TabPFN with simultaneous predictions (``TabPFN (merged)''). The colored bars represent the four different Big Indirect Datasets (BIDs) used as context.}
\label{fig:c1_rmse}
\end{figure*}

The model’s primary advantage was its predictive accuracy. As shown in the RMSE comparison (Fig.\ref{fig:c1_rmse}), TabPFN yielded lower errors across all five boreholes, reducing the RMSE by roughly 20–30\% on average compared to the HBM. The depth profile results (e.g., Fig.\ref{fig:c1_pred_dist_B5}) visually confirm this, showing TabPFN’s predictions tracking the true values more closely than the HBM’s over-smoothed results. Furthermore, the model’s 95\% predictive intervals were well-calibrated, reliably containing the true values and providing trustworthy uncertainty estimates.

TabPFN’s performance was also influenced by the input context, particularly the choice of BID. Consistent with findings from HBM applications, the relevance of the BID content was more critical than sheer volume: more targeted, localized BIDs often outperformed the comprehensive Global-BID/4 (Fig.\ref{fig:c1_rmse}). This principle was further highlighted in the Simultaneous Prediction scenario. By processing all five boreholes together in a single rich context, TabPFN achieved an accuracy nearly identical to the individual-borehole predictions, but with a dramatic improvement in computational efficiency. As detailed in Table\ref{tab:runtime_benchmark1}, the simultaneous run was vastly more efficient than a sequential approach (e.g., 1559 s vs. a cumulative 7685 s for Local-BID/4), demonstrating TabPFN’s strength in handling complex contexts. Our results demonstrate the importance of ``geotechnical prompt engineering,'' i.e., the quality of geotechnical information in the input context significantly affects TabPFN’s performance.


\begin{figure*}[htb]
\centering
\includegraphics[width=\linewidth]{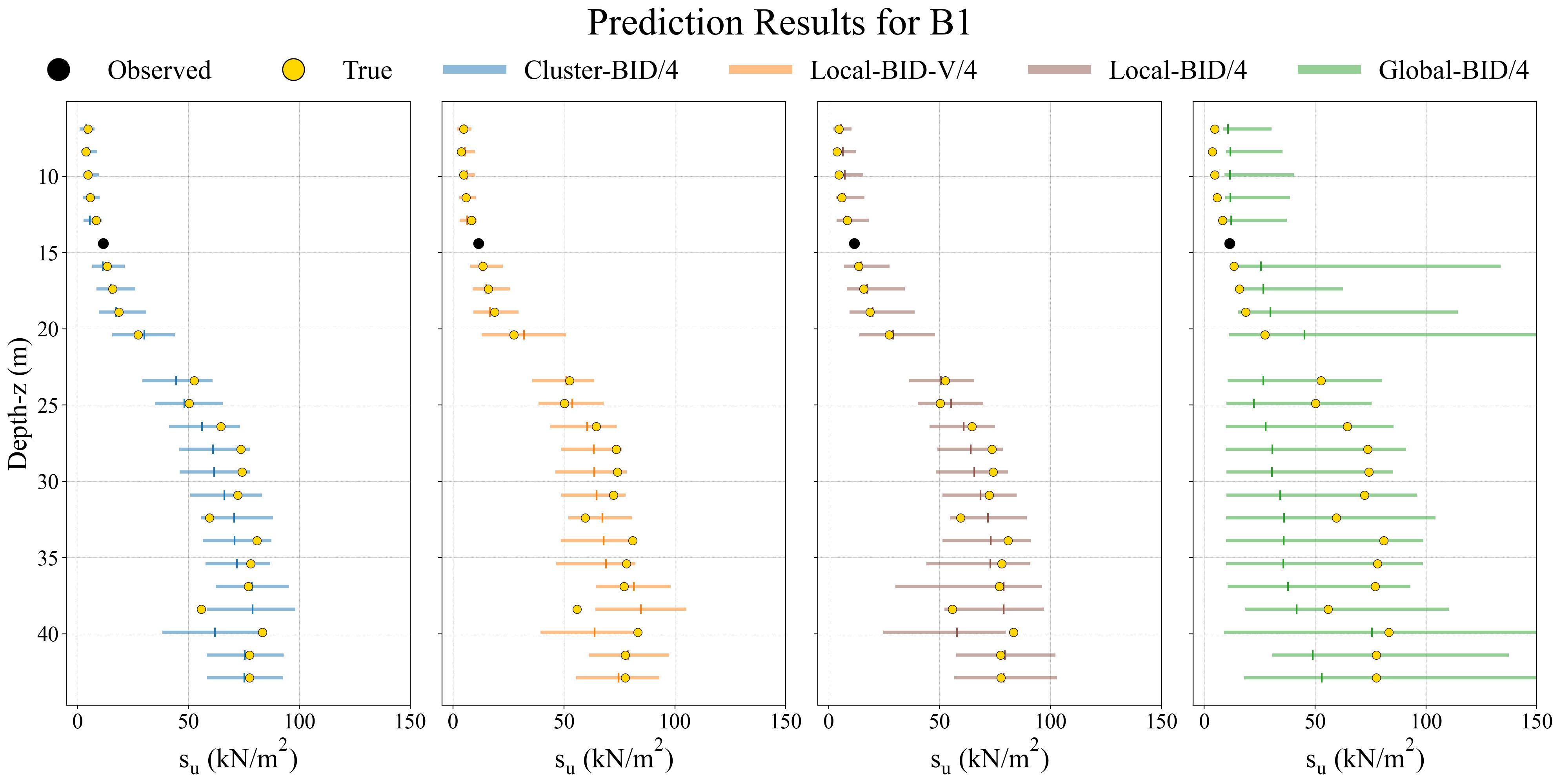}
\caption{Prediction results for the $s_u$ of borehole B1. Each panel shows the TabPFN predictions using a different BID as the predictive context. Observed $s_u$ values provided to the model are shown as black circles. The colored points and horizontal bars represent the mean and the 95\% credible interval of TabPFN's posterior predictive distribution, respectively.}
\label{fig:c1_pred_dist_B1}
\end{figure*}

\begin{figure*}[htb]
\centering
\includegraphics[width=\linewidth]{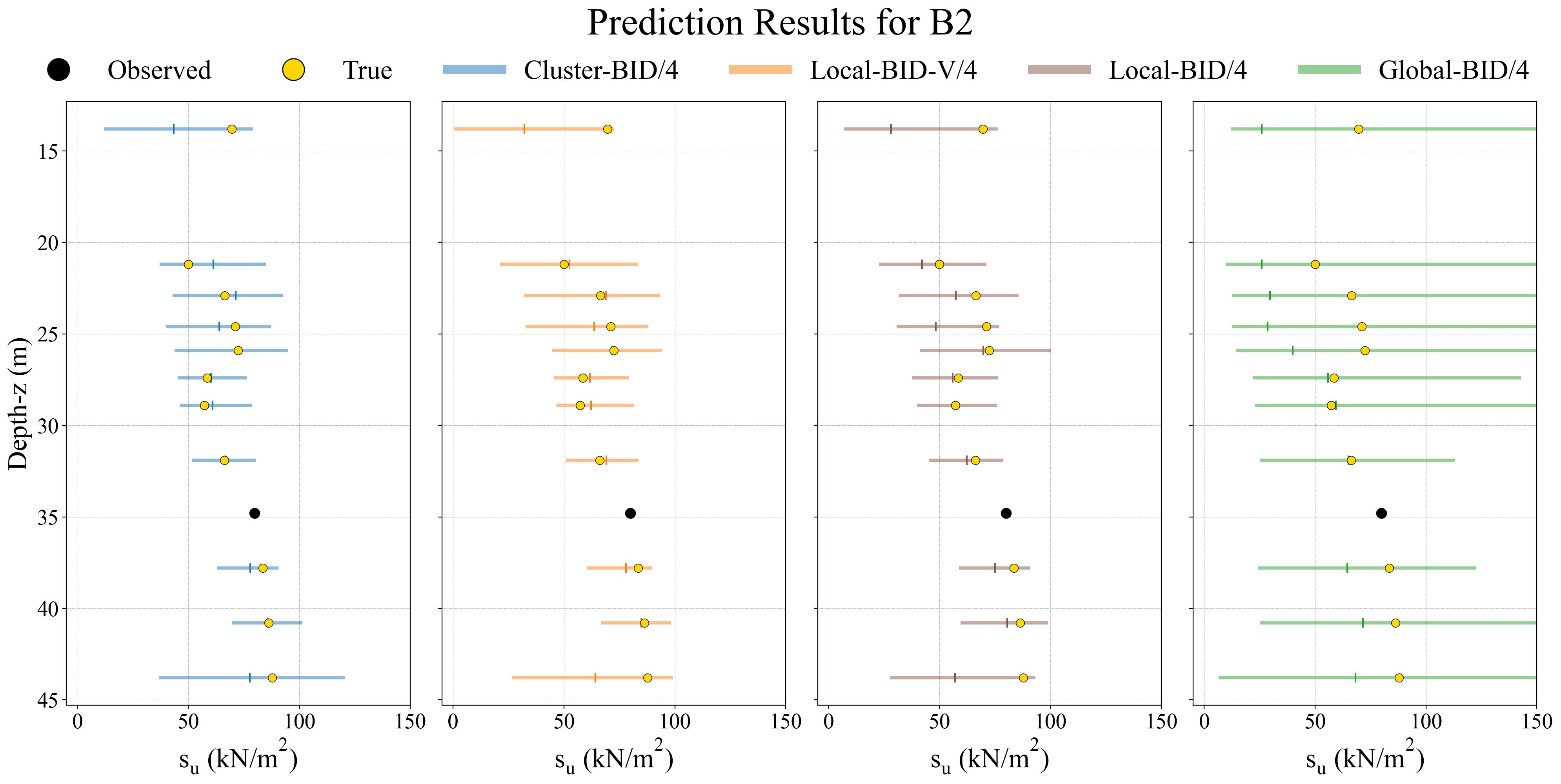}
\caption{Prediction results for the $s_u$ of borehole B2. Each panel shows the TabPFN predictions using a different BID as the predictive context. Observed $s_u$ values provided to the model are shown as black circles. The colored points and horizontal bars represent the mean and the 95\% credible interval of TabPFN's posterior predictive distribution, respectively.}
\label{fig:c1_pred_dist_B2}
\end{figure*}

\begin{figure*}[htb]
\centering
\includegraphics[width=\linewidth]{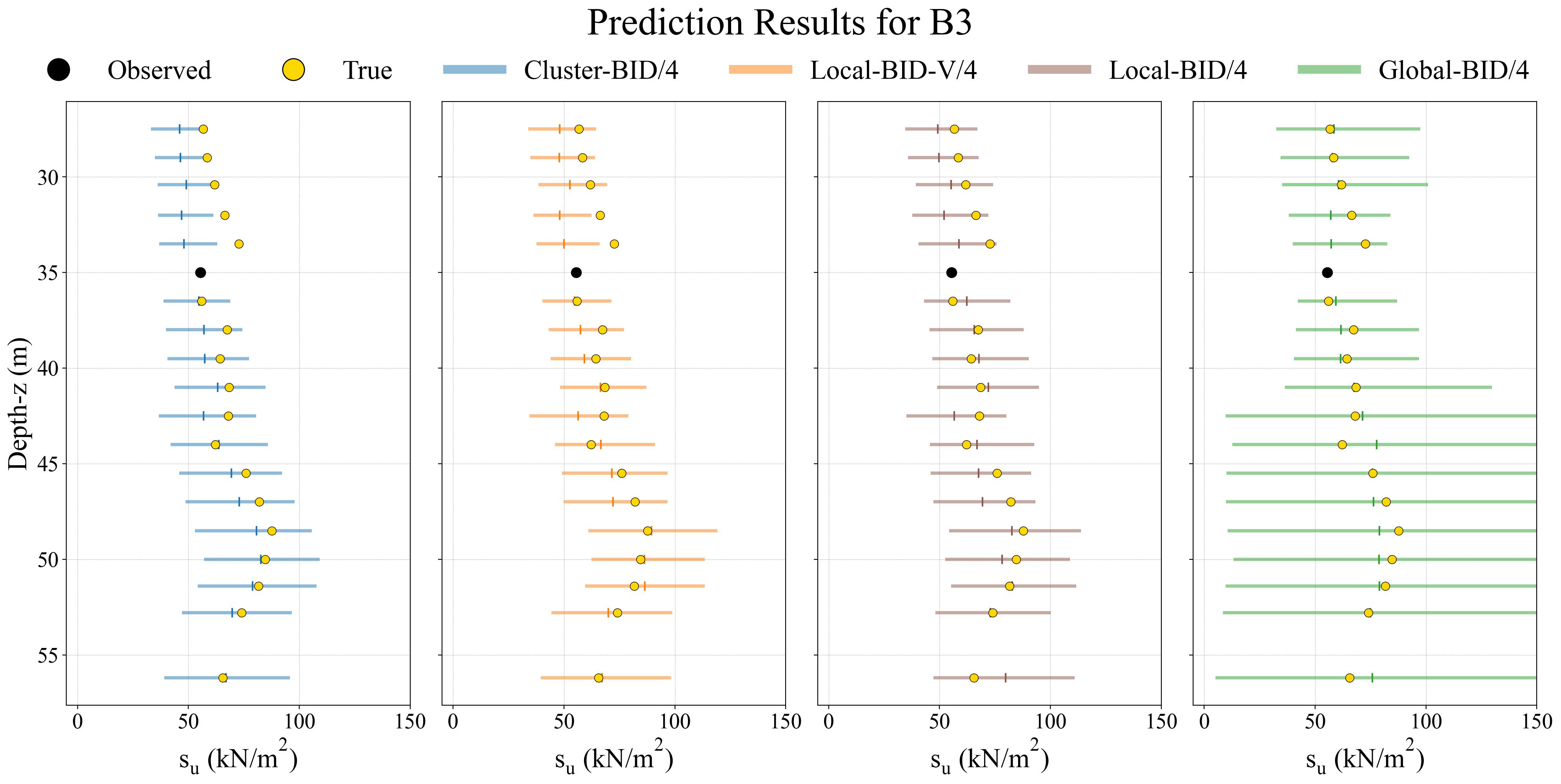}
\caption{Prediction results for the $s_u$ of borehole B3. Each panel shows the TabPFN predictions using a different BID as the predictive context. Observed $s_u$ values provided to the model are shown as black circles. The colored points and horizontal bars represent the mean and the 95\% credible interval of TabPFN's posterior predictive distribution, respectively.}
\label{fig:c1_pred_dist_B3}
\end{figure*}

\begin{figure*}[htb]
\centering
\includegraphics[width=\linewidth]{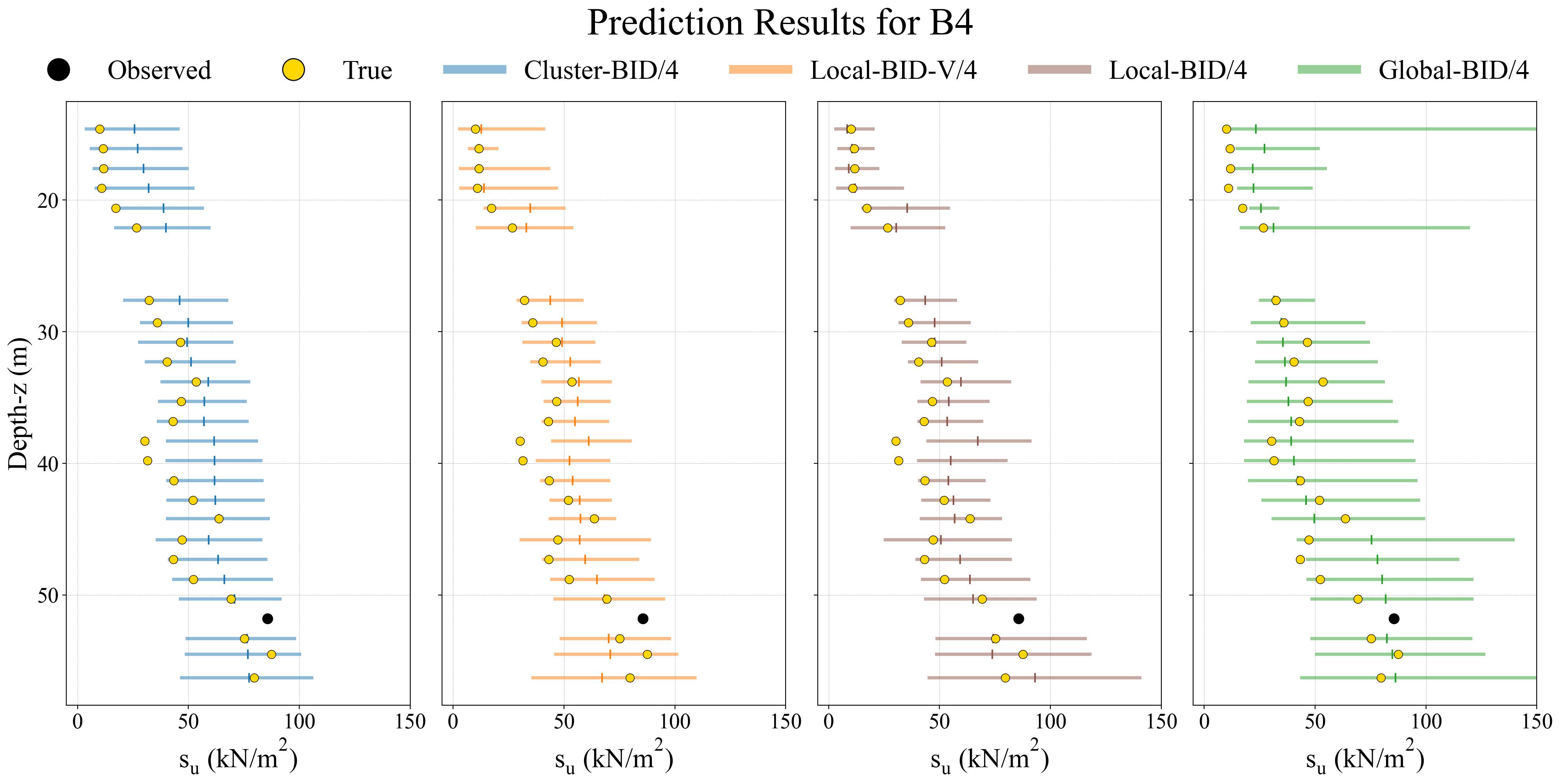}
\caption{Prediction results for the $s_u$ of borehole B4. Each panel shows the TabPFN predictions using a different BID as the predictive context. Observed $s_u$ values provided to the model are shown as black circles. The colored points and horizontal bars represent the mean and the 95\% credible interval of TabPFN's posterior predictive distribution, respectively.}
\label{fig:c1_pred_dist_B4}
\end{figure*}

\begin{table*}[htb]
\centering
\caption{Computational runtimes (seconds) for Benchmark Problem \#1.}
\label{tab:runtime_benchmark1}
\begin{tabular}{l c cc}
\toprule
& & \multicolumn{2}{c}{\textbf{TabPFN Scenarios}} \\
\cmidrule(lr){3-4}
\textbf{BID} & \textbf{HBM (Otake et al., 2025\cite{OTAKE2025100012})}\textsuperscript{a,b} & \textbf{Individual Prediction}\textsuperscript{b} & \textbf{Simultaneous Prediction}\textsuperscript{c} \\
\midrule
Cluster-BID/4 & 121\textsuperscript{d} & 14\textsuperscript{d} & N/A \\
Local-BID-V/4 & 287 & 50 & 53 \\
Local-BID/4   & 2510 & 1537 & 1559 \\
Global-BID/4  & 985 & 85 & 86 \\
\bottomrule
\multicolumn{4}{p{0.9\linewidth}}{\textsuperscript{a} \footnotesize{HBM: Combined runtime for model training on the BID and subsequent inference.}} \\
\multicolumn{4}{p{0.9\linewidth}}{\textsuperscript{b} \footnotesize{Individual Prediction: Average runtime per borehole (averaged over 5 boreholes).}} \\
\multicolumn{4}{p{0.9\linewidth}}{\textsuperscript{c} \footnotesize{Simultaneous Prediction: Total runtime for a single run predicting all 5 boreholes at once.}} \\
\multicolumn{4}{p{0.9\linewidth}}{\textsuperscript{d} \footnotesize{For Cluster-BID/4, values are the average of the five individual clusters (C1--C5).}} \\
\end{tabular}
\end{table*}

\subsection{Benchmark Problem \#2: Predicting missing mechanical parameters}

In the missing-parameter imputation task of Benchmark \#2, the results revealed a clear trade-off: TabPFN achieved vastly superior predictive accuracy, while the HBM offered greater computational efficiency for this multi-target problem.

\begin{figure*}[htb]
\centering
\includegraphics[height=0.9\textheight,keepaspectratio]{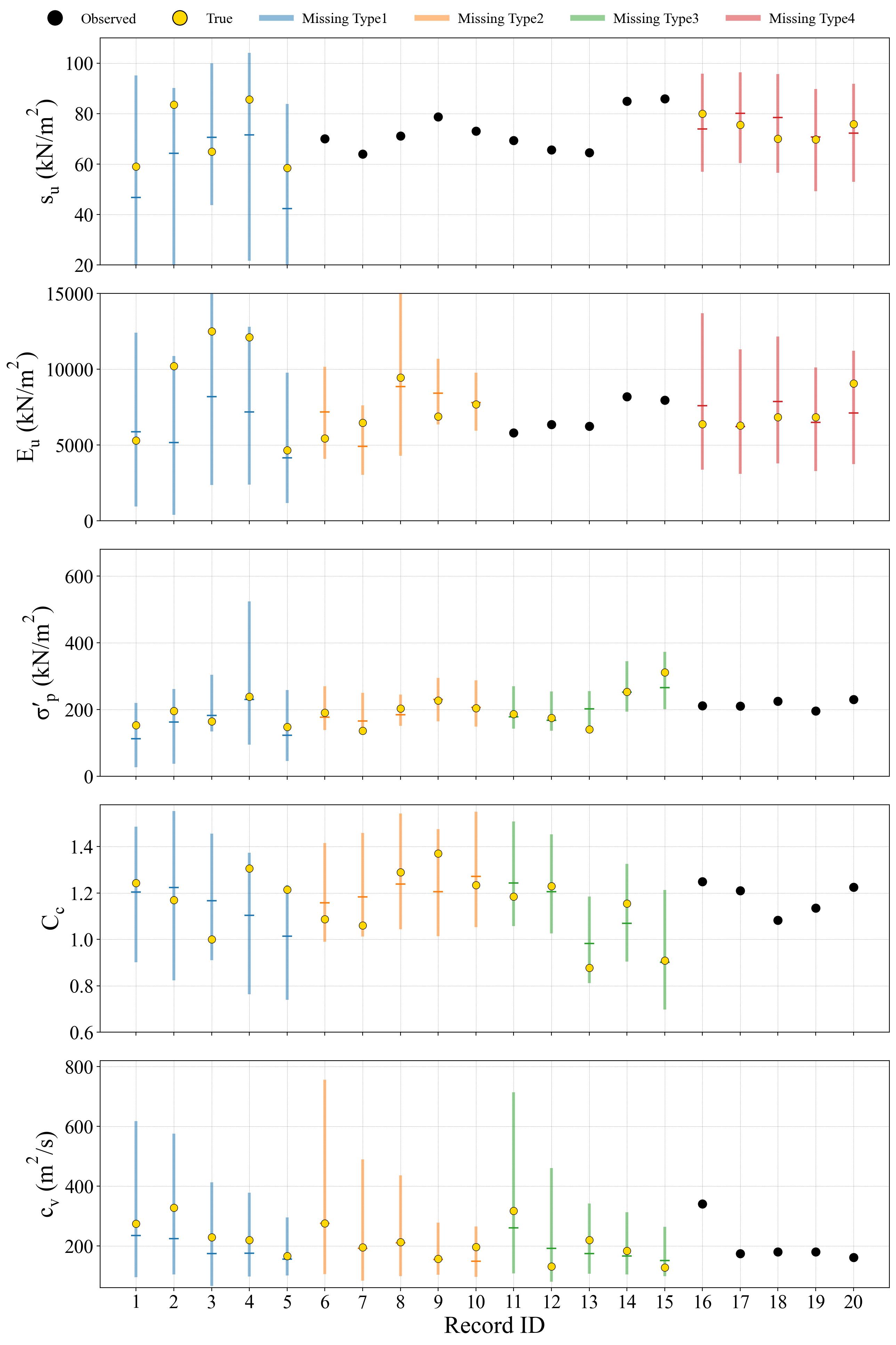} 
\caption{TabPFN estimation results for Benchmark problem \#2}
\label{fig:c2_pred_dist}
\end{figure*}

\begin{figure*}[htb]
\centering
\includegraphics[width=\linewidth]{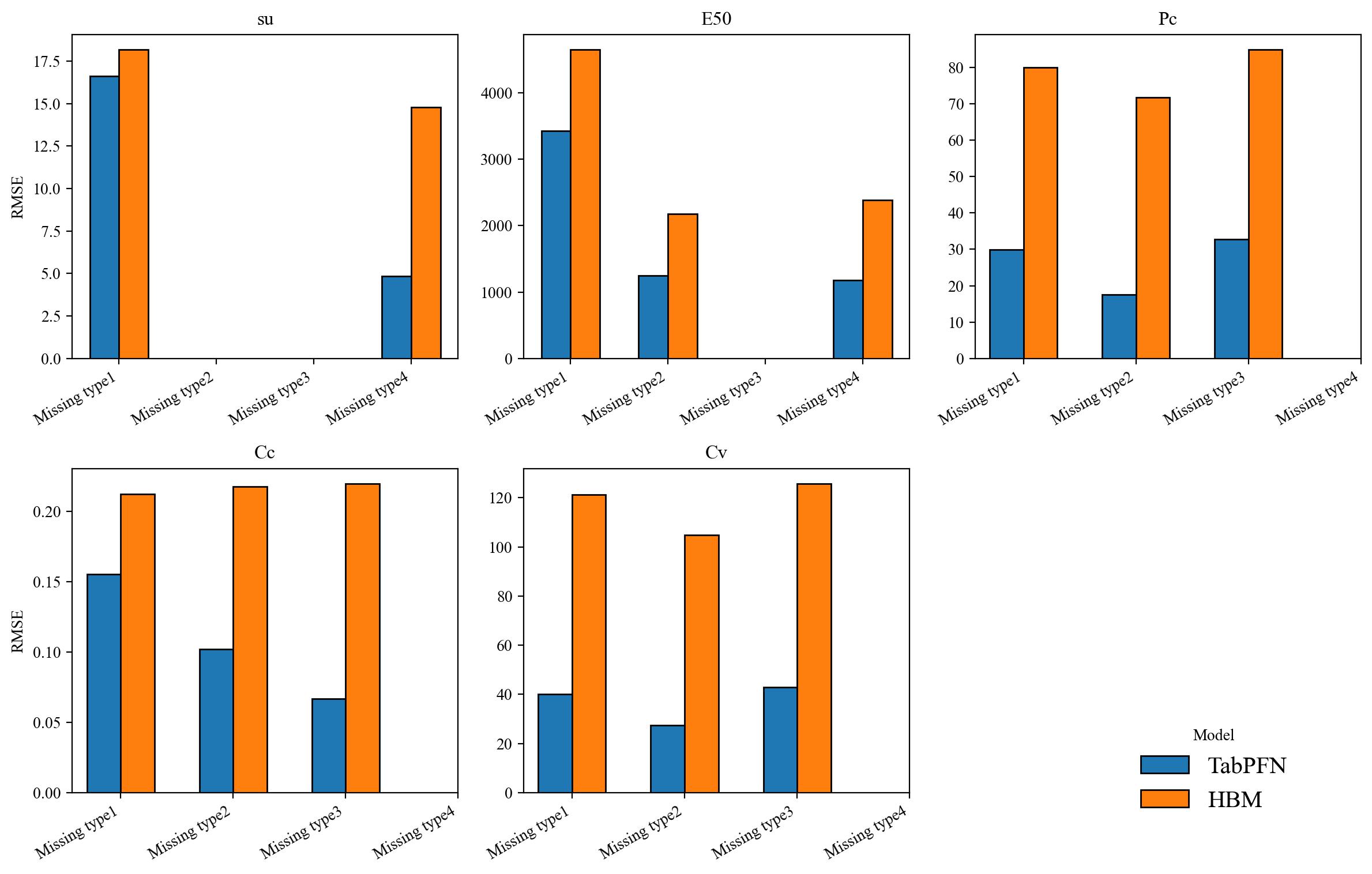}
\caption{Summary of the RMSE results for the mechanical parameters: TabPFN versus sample solutions (Benchmark Problem \#2)}
\label{fig:c2_rmse}
\end{figure*}

TabPFN’s primary advantage was its remarkable accuracy in imputing the five different mechanical parameters. As shown in the RMSE comparisons (Fig.\ref{fig:c2_rmse}), TabPFN consistently and significantly outperformed the HBM across all parameter types and missingness patterns. The imputation results (Fig.~\ref{fig:c2_pred_dist}) visually confirm this, with TabPFN’s estimates (and their uncertainty intervals) aligning closely with the true withheld values.

In terms of computational cost, however, the HBM was more efficient. As detailed in Table\ref{tab:runtime_benchmark2}, the HBM completed the entire imputation task in 452 seconds, whereas the total runtime for TabPFN was 2923 seconds. This difference arises from the models’ fundamental design for this task: the HBM performs multi-parameter imputation within a single integrated model, whereas our use of TabPFN required sequentially running 14 separate models — one for each parameter and missingness pattern — to complete the full task. This highlights a key consideration for multi-target imputation: while TabPFN’s accuracy is superior, the current one-target-at-a-time implementation can lead to a higher cumulative runtime.

\begin{table*}[htb] 
\centering
\caption{Computational runtimes (seconds) for Benchmark \#2: Missing Parameter Imputation.}
\label{tab:runtime_benchmark2}
\begin{tabular*}{\linewidth}{@{\extracolsep{\fill}} l c p{0.6\linewidth} @{}}
    \toprule
    \textbf{Model} & \textbf{Runtime (sec)} & \textbf{Notes} \\
    \midrule
    HBM (Otake et al., 2025\cite{OTAKE2025100012}) & 452 & Runtime for the single HBM to perform the entire imputation task. \\
    \addlinespace 
    TabPFN & 2923 & Total runtime, representing the sum of runtimes from all 14 individual models required to impute all missing parameters. \\
    \bottomrule
\end{tabular*}
\end{table*}

\section{Discussion}

Our findings indicate that TabPFN, a general-purpose foundation model, consistently outperformed a specialized HBM. This section discusses the interpretation and broader implications of these results.

First, TabPFN’s superior accuracy likely stems from its ability to automatically learn complex, non-linear relationships directly from data. The HBM, while effective, relies on pre-specified correlation structures and can struggle when those assumptions do not hold. Furthermore, optimizing an HBM requires significant domain expertise and incurs substantial manual effort in model selection and tuning, whereas TabPFN’s automated, data-driven approach simplifies this process considerably.

Second, our results highlight the emerging concept of ``geotechnical prompt engineering.'' The finding that targeted, localized BIDs often outperform a larger, more general global dataset is crucial. This does not dismiss existing domain knowledge; rather, it reframes it. Principles from established geotechnical data-filtering techniques—where data is selected based on geological similarity (e.g. cluster-based site grouping \cite{WU2022102253, Sharma2022, SHARMA2023105624, CAI2024107537, CAI2025107072}) and methods to capture high-order dependencies among soil properties \cite{SAITO2025100009}—are directly applicable to selecting the most informative context data for TabPFN. This suggests a powerful synergy between traditional geotechnical wisdom and the new data-centric paradigm.

Finally, the practical advantages of TabPFN—its speed and ease of use—signal a potential democratization of advanced probabilistic analysis. By lowering the technical barrier and shifting the heavy computational burden to a one-time pre-training, such models empower practitioners to perform sophisticated analyses more routinely and efficiently, opening the door to real-time applications.

\section{Conclusion and Future Outlook}

In this study, we presented the first benchmark application of TabPFN, a tabular foundation model, to geotechnical site characterization. We demonstrated that on tasks of spatial $s_u$ prediction and parameter imputation, TabPFN offers a powerful, efficient, and accessible alternative to conventional specialized models. Our key findings are:

\begin{itemize}
    \item \textbf{Superior Predictive Performance:} TabPFN substantially outperformed a specialized HBM, achieving up to a 35\% reduction in RMSE and consistently lower errors across all tasks.
    
    \item \textbf{Well-Calibrated Probabilistic Output:} The model generated reliable predictive distributions, providing trustworthy uncertainty estimates essential for risk-informed design, all achieved without relying on strong prior assumptions.

    \item \textbf{Exceptional Efficiency and Usability:} TabPFN’s few-shot inference was an order of magnitude faster than the HBM. Its ``out-of-the-box'' nature, requiring no hyperparameter tuning, significantly lowers the barrier for adopting advanced ML methods in practice.
\end{itemize} 

Looking ahead, several promising research avenues emerge. First, enhancement of TabPFN itself is warranted – particularly by developing multi-output capabilities to improve efficiency in multi-parameter imputation tasks, and by exploring methods to extend its fixed context size. Second, investigating hybrid approaches that combine the data-driven power of TabPFN with the rigor of physics-based models is a key direction; for example, using TabPFN’s outputs as informative priors for an HBM. Finally, the exploration of other foundation models (such as LLMs for integrating textual site investigation reports) will continue to drive the field toward more data-informed and automated analysis.

\section*{Acknowledgement}
This research was supported by JSPS KAKENHI Grant Number JP23H00195 and the ``Strategic Research Projects site'' grant from ROIS (Research Organization of Information and Systems). We would like to express our gratitude for the support.

\bibliographystyle{junsrt}
\bibliography{refs}

\end{document}